\documentclass[conference]{IEEEtran}
\usepackage{graphicx}
\usepackage{subfigure}
\usepackage{wrapfig}
\usepackage{cite}
\usepackage{multirow}

\begin{document}

\title{Deep Learning Object Detection Methods for Ecological Camera Trap Data}

\author{
    \IEEEauthorblockN{Stefan Schneider\IEEEauthorrefmark{1}, Graham W.~Taylor\IEEEauthorrefmark{2}, Stefan C.~Kremer\IEEEauthorrefmark{1}}
    \IEEEauthorblockA{\IEEEauthorrefmark{1}School of Computer Science,
		University of Guelph\\
    \{sschne01, skremer\}@uoguelph.ca}
		\IEEEauthorblockA{\IEEEauthorrefmark{2}School of Engineering,
		University of Guelph\\
    gwtaylor@uoguelph.ca}
    \IEEEauthorblockA{\IEEEauthorrefmark{2}Vector Institute for Artificial Intelligence
    }
		\IEEEauthorblockA{\IEEEauthorrefmark{2}Canadian Institute for Advanced Research
    }
}

\maketitle

\begin{abstract}
Deep learning methods for computer vision tasks show promise for automating the data analysis of camera trap images. Ecological camera traps are a common approach for monitoring an ecosystem's animal population, as they provide continual insight into an environment without being intrusive. However, the analysis of camera trap images is expensive, labour intensive, and time consuming. Recent advances in the field of deep learning for object detection show promise towards automating the analysis of camera trap images. Here, we demonstrate their capabilities by training and comparing two deep learning object detection classifiers, Faster R-CNN and YOLO v2.0, to identify, quantify, and localize animal species within camera trap images using the Reconyx Camera Trap and the self-labeled Gold Standard Snapshot Serengeti data sets. When trained on large labeled datasets, object recognition methods have shown success. We demonstrate their use, in the context of realistically sized ecological data sets, by testing if object detection methods are applicable for ecological research scenarios when utilizing transfer learning. Faster R-CNN outperformed YOLO v2.0 with average accuracies of 93.0\% and 76.7\% on the two data sets, respectively. Our findings show promising steps towards the automation of the labourious task of labeling camera trap images, which can be used to improve our understanding of the population dynamics of ecosystems across the planet.
\newline
\end{abstract}


\section{Introduction}
\noindent
Population ecologists use camera traps to monitor animal population sizes and manage ecosystems around the world. Camera traps were first introduced in 1956, and in 1995, Karanth demonstrated their usefulness for population ecology by re-identifying tigers (\textit{Panthera tigris}) in Nagarahole, India using a formal mark and recapture model \cite{gysel1956simple, karanth1995estimating}. The popularity of the camera trap methodology grew rapidly thereafter, with a 50\% annual growth using the technique as a tool to estimate population sizes \cite{burton2015wildlife, rowcliffe2008surveys}. Camera traps respond to motion, which generally corresponds with an animal entering the frame. Camera trap data analyses involve manually quantifying the species and number of individuals in thousands of images. Automating this process has obvious advantages, including a reduction in human labour, an unbiased estimate across analyses, and the availability of species identification without domain expertise.
\newline
\\
In this work, we focus on utilizing deep learning based approaches for object detection to identify, quantify, and localize animal species within camera trap images. Camera trap data provides a robust measure of the capabilities of deep learning for species classification, as the images are often `messy', with animals being partly obstructed, positioned at varying distances, cropped out of the image, or extremely close to the camera \cite{norouzzadehautomatically}. These obstacles are in addition to the traditional difficulties of computer vision tasks, such as variable lighting, photos taken at day and night, and species exhibiting a variety of poses.
\newline
\\
Deep learning methods have demonstrated near perfect accuracy for computer vision tasks when trained on large labled datasets; however, labeled ecological data is notorious for being sparse and intermittent \cite{chao1989estimating}. We aim to test the bounds of deep learning for realistic ecological applications, demonstrating the usefulness of the technique for researchers to train their own classifiers on their own ecosystem of interest, instead of relying on large public data sets which may not fit their niche of study. We considered the Reconyx Camera Trap data set, which contains 946 labeled images with 20 species classifications and bounding box coordinates, as well as the Gold Standard Snapshot Serengeti data set, which contains 4,096 labeled images of 48 species classifications \cite{yu2013automated, norouzzadehautomatically}. Current methods for object detection require the bounding box coordinates for training, and as a result, we hand-labeled the bounding box coordinates for the Gold Standard Snapshot Serengeti data set and offer it to the camera trap and deep learning community.
\newline
\\
We compare two methods for object detection using deep learning, Faster Region-Convolutional Neural Network and You-Only-Look-Once v2.0 (hereafter referred to as Faster R-CNN and YOLO, respectively) \cite{redmon2016you, ren2017faster}. These two approaches are generally considered by the trade-off of data efficiency versus speed, as YOLO can be used in real time, but requires additional training data \cite{redmon2016you}. Our results demonstrate Faster R-CNN shows promise for accurate and autonomous analysis of camera trap data, while YOLO fails to perform. These results demonstrate that ecologists should consider utilizing Faster R-CNN or its successors as the method of object detection to autonomously extract ecological information from camera trap images.

\section{Background and Related Work}

\noindent
\textbf{Deep Learning for Object Detection:} Many recent advancements in deep learning have come from improving the architectures of a neural network. One such architecture is the Convolutional Neural Network (CNN), which is now the most commonly used architecture for computer vision tasks \cite{fukushima1979neural, krizhevsky2012imagenet}. CNNs introduce convolutional layers within a network which, for a given image, learn many feature maps which represent the spatial similarity of patterns found within the image (such as colour clusters, or the presence or absence of lines)~\cite{lecun2015deep}. Each feature map is governed by a set of `filter banks', which are matrices of scalar values that can generally be considered synonymous to the weights of a feedforward network. For each convolutional layer, the filter banks are similarly passed through a non-linear transformation and learned using gradient descent with backpropagation \cite{lecun2015deep}. CNNs also introduce max pooling layers, a method that reduces computation and increases robustness by evenly dividing the feature map into regions and returning only the highest activation values \cite{lecun2015deep}. As a result of having numerous feature maps for a given input, CNNs are particularly well suited for dealing with data from multiple arrays, such as colour images, which have three colour channels \cite{lecun2015deep}. Deep learning researchers continually experiment with the modular architectures of neural networks and four CNN frameworks have been standardized as well-performing with differences including computation cost and memory in comparison to accuracy. These networks include AlexNet, VGG, GoogLeNet/InceptionNet (which introduced the inception module), and ResNet, which introduced skip connections \cite{krizhevsky2012imagenet, jaderberg2015spatial, szegedy2015going, he2016deep}. These networks range from 7 to 152 layers. A common approach to training deep learning classification tasks is to use publicly available weights from one of these four network structures trained on a public data set as initialization parameters, and retraining the network using your own limited data set \cite{pan2010survey}. This allows for learned filters, such as edge or colour detectors, to be used without having to be re-learned on limited data. This technique is known as Transfer Learning \cite{pan2010survey}.
\newline
\\
CNNs have demonstrated great success for image classification, conditioned on the network being trained to return a single label for a given image \cite{krizhevsky2012imagenet}. In order to determine the classification of more than one object within an image, computer vision researchers train an object detector, where the image is segregated into overlapping regions (often called `proposals') \cite{girshick2015fast}. Two approaches for object detection have seen wide-spread success. The earliest approach was R-CNN, where an image is crudely segregate into a series of different sized boxes using an image segregation algorithm, and each region is passed through a CNN. Fast R-CNN introduced region proposals generated based on the refined last feature map of the network to decrease proposal computation \cite{girshick2015fast}. Soon after, Faster R-CNN, which introduces a Region Proposal Network (RPN) to the framework, enabled nearly cost-free region proposals \cite{he2016deep}. A second approach for object detection is YOLO, which divides an image into a grid, with each gridcell acting as the origin for numerous predefined `anchors' relevant to the size classifications of interest. For example, when searching for a cat, one may implement three anchors: a square, a horizontal rectangle, and a vertical rectangle, as a cat may approximately fit into each shape. When training and using YOLO, output classifications are returned for every anchor in a single iteration. \cite{redmon2016you}. YOLO is often less accurate due to the static nature of the anchor boxes, but has been shown to be 3x faster than Faster R-CNN~\cite{redmon2016you}.
\newline
\\
\noindent
\textbf{Automating the Analysis of Camera Trap Images:}
Prior to the wide-spread adoption of deep learning systems, computer vision researchers developed a variety of creative and moderately successful methodologies for the automated analysis of animals from camera traps based on the raw pixel data from images. Initial approaches for species classification required a domain expert to identify meaningful features for the desired classification (such as the unique characteristics of animal species), design an algorithm to extract these features from the image, and compare individual differences using a statistical analysis. Computer vision systems were first introduced for species classification within the microbial and zooplankton community to help standardized species classification, and considered morphological silhouettes \cite{jeffries1984automated, simpson1991classification, balfoort1992automatic}. The first complete camera trap analysis was done in 2013 using the Scale-Invariant Feature Transformation algorithm in combination with a Support Vector Machine to classify species using the Reconyx Camera Trap data set after a foreground extraction technique was applied to separate the animal from the background \cite{lowe2004distinctive, suykens1999least, yu2013automated}.
\newline
\\
In 2014, Chen et al. \cite{chen2014deep} reported the first paper for animal species classification using a CNN that considered the Reconyx Camera Trap data set. Their CNN was a shallow network by modern standards, with 3 convolution and 3 pooling layers.
\newline
\\
In 2016, Gomez et al. \cite{gomez2016towards} used deep CNNs for camera trap species recognition, comparing 8 variations of the established CNN frameworks AlexNet, VGG, GoogLeNet, and ResNet to train species classification on the complete Snapshot Serengeti data set of 3.2 million images with 48 species classifications. The ResNet-101 architecture achieved the best performance. Following this work, they also utilized deep learning to improve low resolution animal species recognition by training deep CNNs on poor quality images. The data was labeled by experts into two data sets, the first classifying between birds and mammals and the second classification of different mammal species \cite{gomez2016animal, caruana1998multitask}.
\newline
\\
In 2017, Norouzzadeh et al. \cite{norouzzadehautomatically}, utilized the ability of a network to return numerous output classifications for a given image, a technique known as multitask learning, to consider the species, quantify the number of animals, as well as to determine additional attributes. This approach operates differently than object detection methods, as their classifier learns what an image with a given number of animals looks like, rather than individually detecting the number of individuals within the image. Nine independent architectures were trained, including AlexNet, VGG, GoogLeNet, and numerous variations of ResNet. The authors report a species classification accuracy, counting, and attribute accuracy considering an ensemble of their nine models \cite{norouzzadehautomatically}.
\newline
\\
These approaches all share the common limitation of returning only one output per classification task per image, which is unrealistic for meaningful camera trap data analyses. Object detection methods account for this limitation, allowing for a classifier to return multiple species as output.
\newline
\\
\section{Experiments and Results}
\noindent
\textbf{Reconyx Camera Trap data set and Snapshot Serengeti Project:} The Reconyx Camera Trap (further referred to as RCT) data set is a collection of 7,193 camera trap images from two locations in Panama and the Netherlands, capturing colour images during the day, and gray-scale at night \cite{yu2013automated}. Of all the images, only a subset of 946 images include labeled bounding box coordinates, and so we only considered these images.
\newline
\\
The Snapshot Serengeti data set is the world's largest publicly available collection of camera trap images, with approximately 1.2 million images collected using 225 camera traps since 2011 \cite{swanson2015snapshot}. To provide labels, the organization has created a website where nearly 70,000 individuals help label the images by selecting predefined classifications of the species, the number of individuals (1, 2, 3, 4, 5, 6, 7, 8, 9, 10, 11-50, 50+), various behaviours (i.e.,~standing, resting, moving, eating, or interacting), and the presence of young. In additional, there is the Gold Standard Snapshot Serengeti (further referred to as GSSS) data set which contains 4,432 images labeled by experts within the field; however only classification and not bounding box co-ordinates. We annotated the GSSS data set to test object detection methods and give these to the Snapshot Serengeti community. The labeled data set is available at: \textit{https://dataverse.scholarsportal.info/dataset.xhtml?persistentId =doi:10.5683/SP/TPB5ID}.
\newline
\\
For this experiment, we consider the ResNet-101 architecture for both object detection methods. ResNet-101 is a robust network that showed great success in other camera trap studies \cite{gomez2016towards}. We initialized both object detection classifiers using a pre-trained model of the Common Object in Context 2017 data set \cite{lin2014microsoft}. The weights of the final layer were initialized using the Xavier initialization \cite{he2015delving}. Each model was trained using the adaptive momentum optimizer, and training concluded after the loss failed to improve after 3 successive epochs \cite{kingma2014adam}.
\newline
\\
For both data sets, the bounding box coordinates only pertain to a subset of the larger data set. As a result, there has been no prior experimentation and division of standardized train/test labels. In order to account for this, we perform a cross-validation-based evaluation, repeating the procedure five times and reporting the mean and standard deviation across these runs considering a 80/20 train/test split. To improve accuracy, bounding boxes containing less than 750 pixels were removed from the data set.
\newline
\\
We consider two performance metrics, accuracy and Intersection Over Union (IOU). Accuracy represents the percentage of correctly classified species. IOU is an evaluation metric specific to the performance of object detection methods. IOU returns performance as the area of overlap of the true and predicted regions divided by the entire area of the true and predicted regions \cite{nowozin2014optimal}. To quantify accuracy using object detection, numerous classification comparisons are calculated per image. To do this, we calculate the IOU for each predicted box for an image in comparison to a test box, select the highest IOU, and then compare its classification output to the true classification. After bounding boxes are used for a classification, they are removed for future comparisons. IOU values above 0.70 are considered well performing \cite{nowozin2014optimal}.
\newline
\\
Faster R-CNN returned an accuracy of 93.0\% and 76.7\%, and IOU values of 0.804 and 0.722 on the RCT data set and GSSS data set, while YOLO returned an accuracy of 73.0\% and 40.3\% and IOU of 0.570 and 0.221, respectively (Table 1). Faster R-CNN returned an accuracy of 100\% on 13 of the 18 species considered in the RCT data set, and 80\% accuracy on 5 of the 11 species considering species with more than 100 images in the GSSS data set (Table 2 \& 3). Figures 1-3 and 4-6 are examples of the Faster R-CNN performance for the RCT and GSSS data set respectively. 
\newline

\begin{table}[h]
\centering
\caption{Comparison of Faster R-CNN and YOLO performance based on accuracy and IOU}
\begin{tabular}{ c c c c }
	\hline
	\textbf{Data Set} & \textbf{Model} & \textbf{Acc. (\%)} & \textbf{IOU}\\
	\hline
	RCT & Faster R-CNN & 93.0 $\pm$ 3.20 & 0.80 $\pm$ 0.03\\
	 &  YOLO & 65.0 $\pm$ 12.1 & 0.57 $\pm$ 0.09\\
	GSSS & Faster R-CNN & 76.7 $\pm$ 8.31 & 0.72 $\pm$ 0.08\\
	 & YOLO & 43.3 $\pm$ 14.5 & 0.22 $\pm$ 0.12\\
	\hline
\end{tabular}
\begin{flushleft}
\end{flushleft}
\end{table}

\section{Discussion}
\noindent
By utilizing modern approaches for object-detection, we demonstrate that researchers that require the analysis of camera trap images can automate animal identification, quantification, and localization within images. Previous studies have demonstrated the quantification of animal individuals from camera trap data, but they suffer the limitation of returning a single classification per image, which is unrealistic for camera trap data. We demonstrate that Faster R-CNN is capable of accurately classifying more than one species per image given limited data when utilizing transfer learning.
\newline
\\
Deep learning has demonstrated super-human performance on tasks with large amounts of data; however we test the reliability of deep learning methods on realistically sized ecological camera trap data sets. Without this distinction, deep learning approaches for autonomous camera trap data analysis would be limited to ecosystems with large numbers of labeled camera trap data, like Snapshot Serengeti, which required the effort of thousands of individuals to label. We demonstrate that if a research group performs a one-time labeling of less than 1,000 images, one can create a reliable model using Faster R-CNN. Our YOLO model performed poorly on both data sets, likely due to limited data.
\newline
\\
While the GSSS data set contained approximately 4x the number of images, the trained model for the data set performed worse than the trained model for the RCT data set using Faster R-CNN. There are numerous explanations for this. First, the GSSS data set has extreme class imbalances, a well documented scenario where machine learning classifiers have had difficulty \cite{kotsiantis2007supervised}. In addition, the GSSS data set is much `messier' than the RCT data set, with the majority of images containing animals either extremely far away, cropped by the camera, obstructed behind another object/animal, and/or extremely close to the camera. While the RCT data set does contain some of these difficult scenarios, there are far fewer occurrences. When implementing models such as these, our results reiterate the importance of class balance. For real-life applications, if an animal of interest rarely appears in the camera trap data, we recommend finding and labeling additional images from outside sources to build a balanced data set, or exploring additional techniques for class imbalance.
\newline
\\
Considering the success of the Faster R-CNN model, our method allows for future possibilities regarding detailed individual and behaviour analysis from camera trap images. Norouzzadeh et al. (2017) demonstrated this in its infancy by returning labeled classifications of young versus adult and male versus female classifications, and the specific behaviour found within the image  \cite{norouzzadehautomatically}. This approach is not reliable, as if more then one species, age, sex, or behaviour are present, the classifier returns erroneous results. Object detection methods allow for the classifier to identify an age, sex, and behaviour of each individual within the image. Using this method of data collection, examples of autonomous ecological reports based on images with time-stamps include: comparing the movement patterns of genders within and across species, identifying seasonally when reproduction occurs by quantifying when infants are most active, and general comparisons of activity/behaviour across species, sex, and age.
\newline
\\
While object detection provides promising steps forward, in order to reliably quantify population metrics, an automated system must be able to re-identify an individual it has previously seen. Camera trap re-identification methods suffer from an unavoidable bias when analyzed by a human and there is debate arguing against the reliability of humans when re-identifying animal individuals from camera trap data \cite{foster2012critique}. The development of a method for reliable animal re-identification would allow for autonomous population estimation of a given habitat using a formal mark and recapture model, such as Lincoln-Petersen \cite{robson1964sample}. Population estimates are reliant on accurate animal identification and if a deep learning system can demonstrate accurate animal re-identification, one could utilize these methodologies to create autonomous systems to extract a variety of ecological metrics, such as diversity, relative abundance distribution, and carrying capacity, contributing to larger overarching ecological interpretations of trophic interactions and population dynamics.
\newline
\section{Conclusion}
\noindent
Recent advancements in the field of computer vision and deep learning have given rise to reliable methods of object detection. We demonstrated the successful training of an object detection classifier using the Faster R-CNN model considering limited ecological camera trap data. Utilizing object detection techniques, ecologists can now autonomously identify, quantify, and localize individual species within camera trap data without the previous limitation of returning only one species classification per image. Our findings show promising steps towards the automation of the labourious task of labeling camera trap images which can be used to improve our understanding of the population dynamics of ecosystems across the planet.

\section*{Acknowledgments}
\noindent
The authors would like to thank all the Snapshot Serengeti volunteers and the camera trap community at large for uploading their data for public access. We also thank the deep learning community at large for their continued pursuit of open sourcing new methodologies encouraging interdisciplinary works.

\bibliographystyle{IEEEtran}
\bibliography{IEEEQuantifyingAnimalCameraTraps}
\twocolumn

\begin{figure}
	\includegraphics[width=8.5cm]{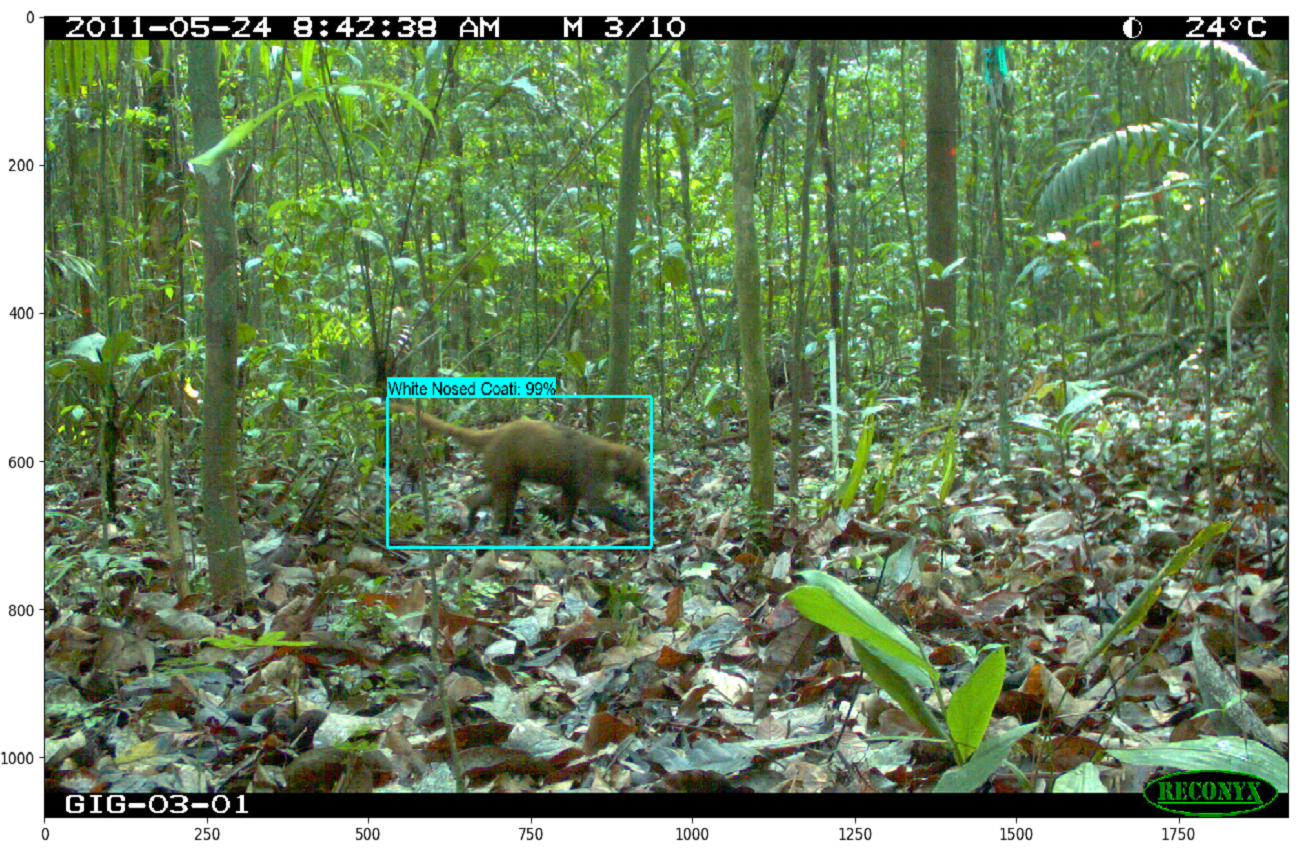}
	\caption{Faster R-CNN output returning 1 White Nosed Agouti from the RCT data set in a highly camouflaged environment.}
\end{figure}

\begin{figure}
	\includegraphics[width=8.5cm]{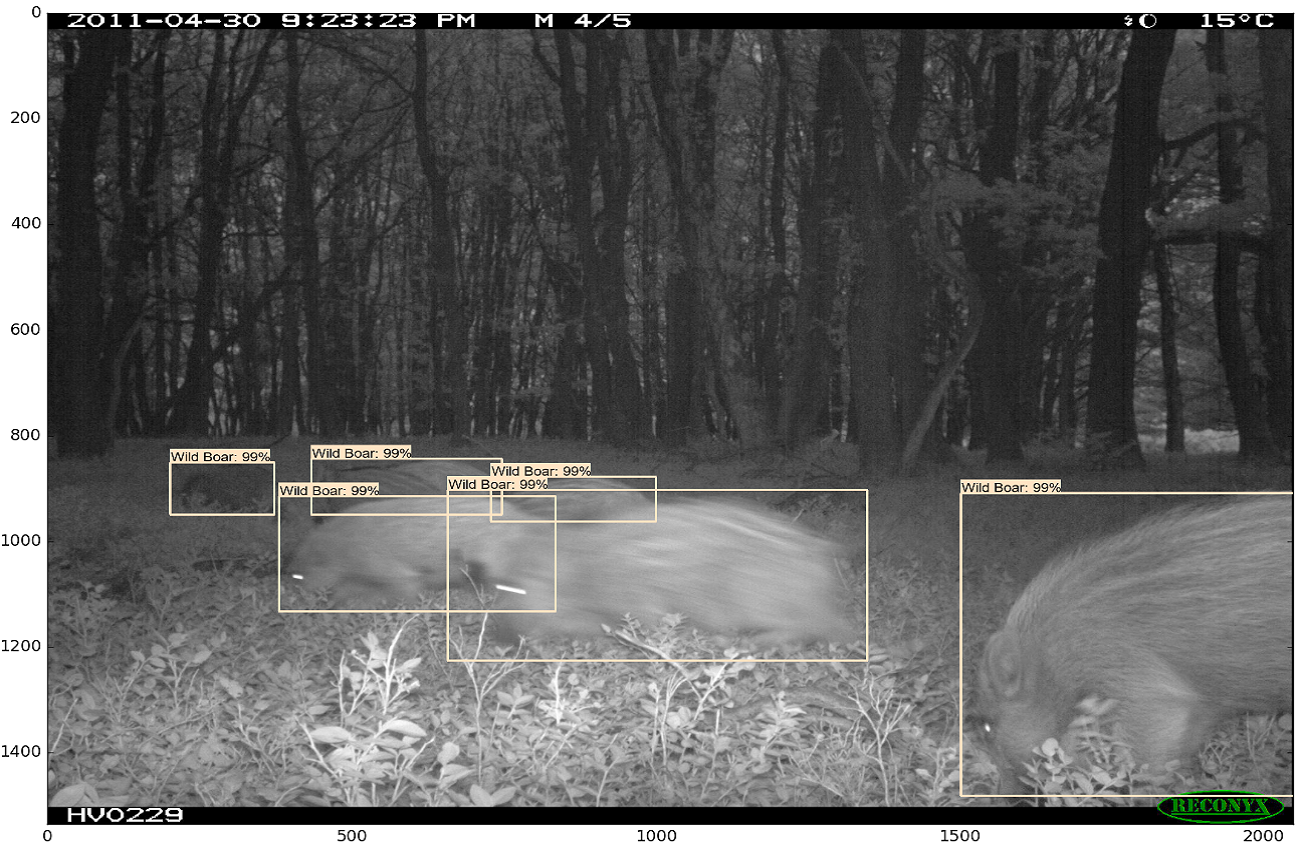}
	\caption{Faster R-CNN output returning 6 wild boar classifications from the RCT data set in an image taken at night.}
\end{figure}

\begin{figure}
	\includegraphics[width=8.5cm]{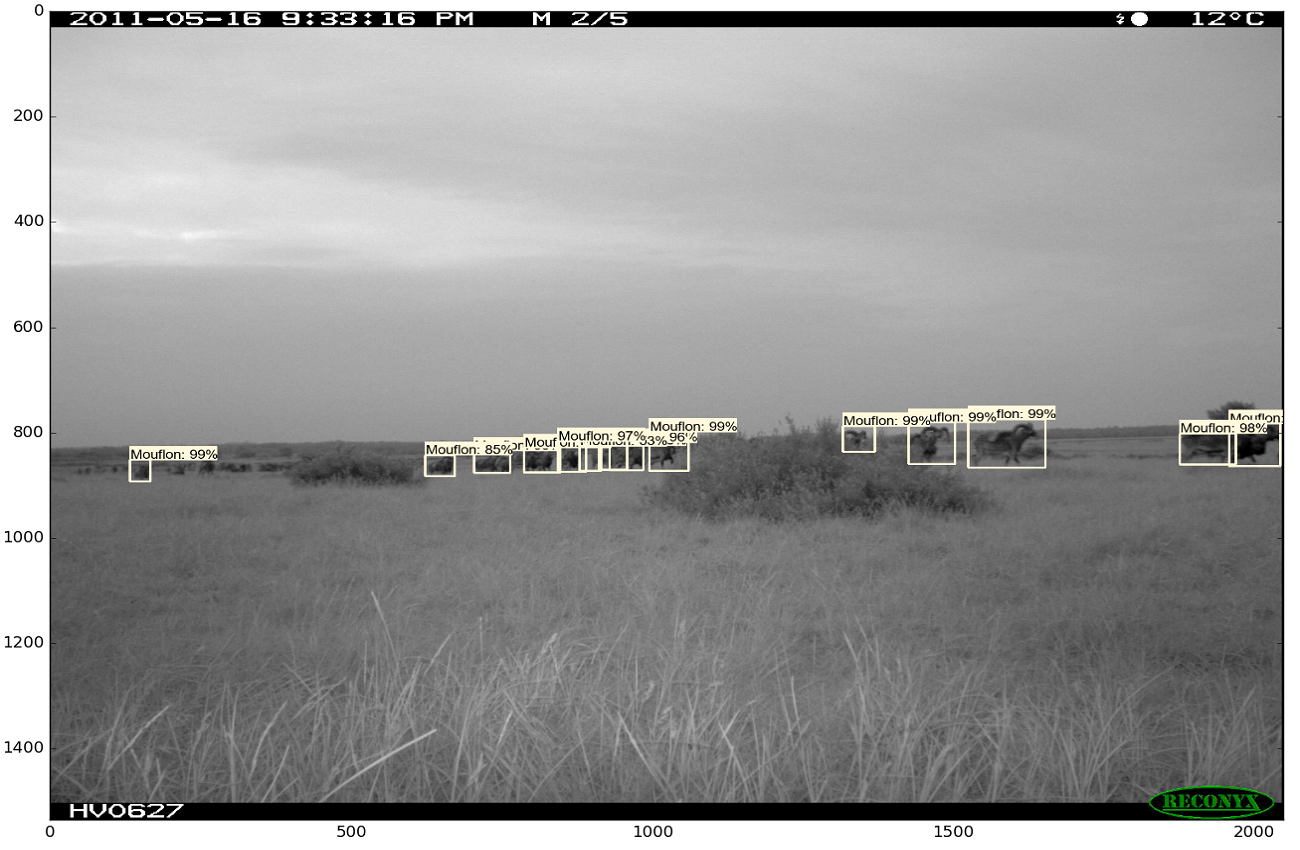}
	\caption{Faster R-CNN output returning 15 Mouflon classifications from the RCT data set in an image taken at night.}
\end{figure}
\newpage

\begin{figure}
	\includegraphics[width=8.5cm]{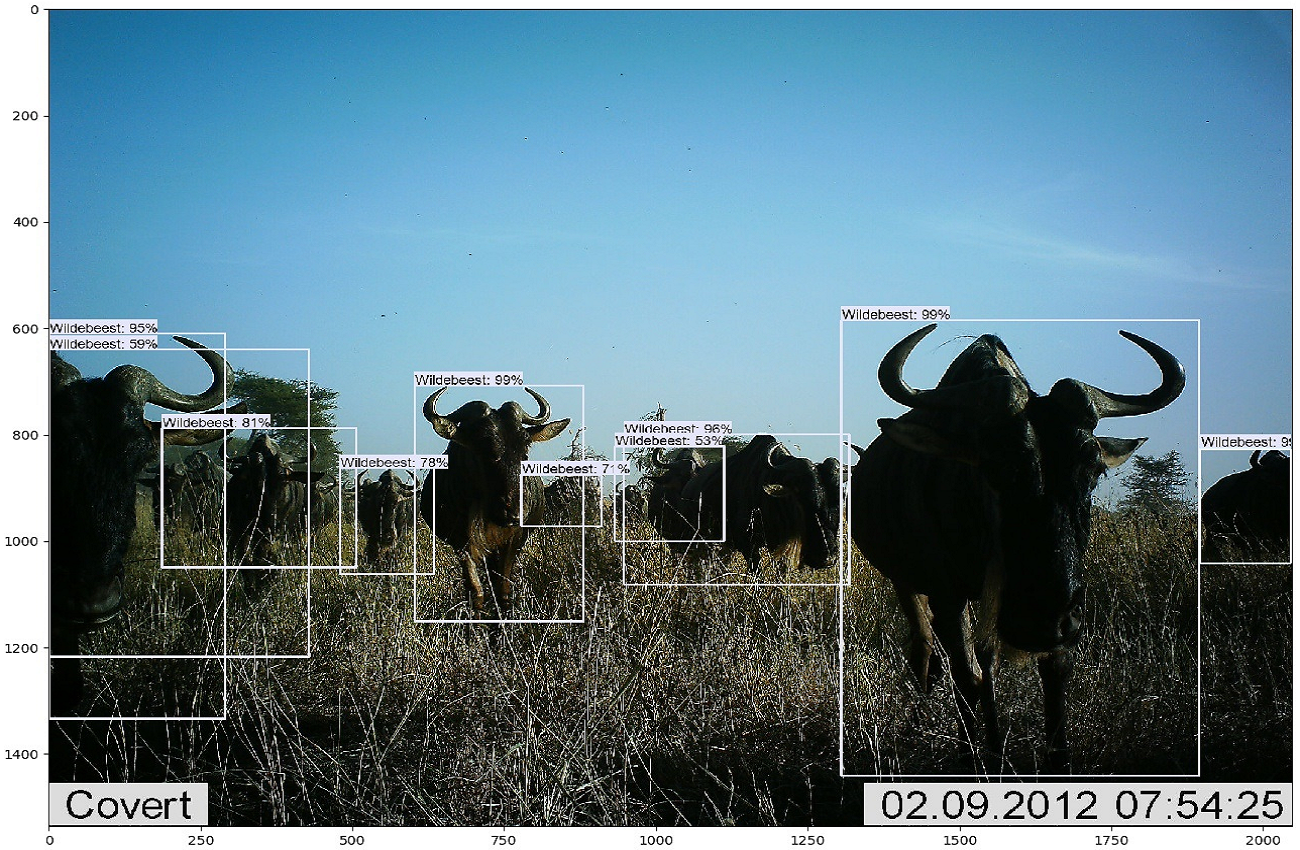}
	\caption{Faster R-CNN output returning 10 Wildebeest from the GSSS data set, demonstrating one example of the high levels of obstruction within the data set.}
\end{figure}

\begin{figure}
	\includegraphics[width=8.5cm]{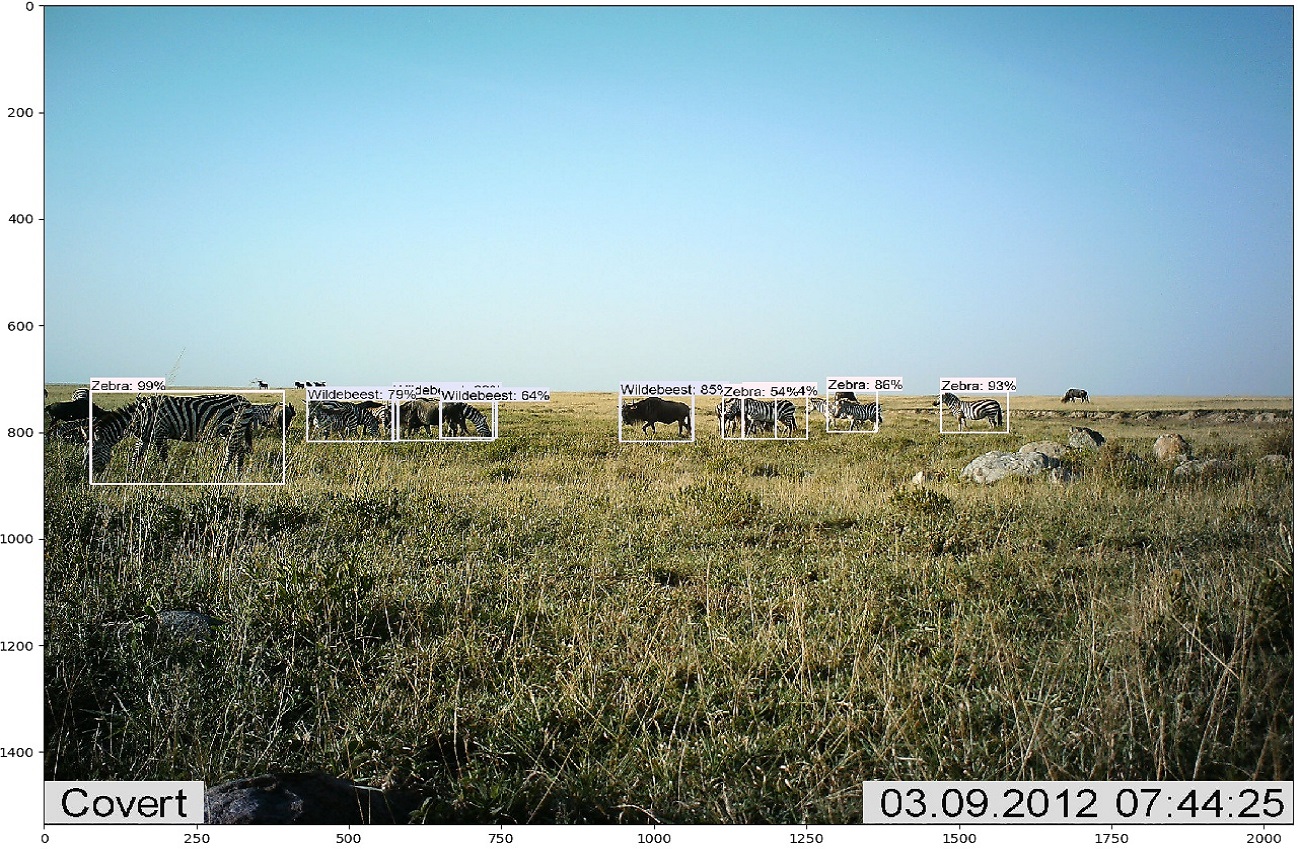}
	\caption{Faster R-CNN output returning 4 Zebra and 4 Wildebeest from the GSSS data set, demonstrating two species within one image.}
\end{figure}

\begin{figure}
	\includegraphics[width=8.5cm]{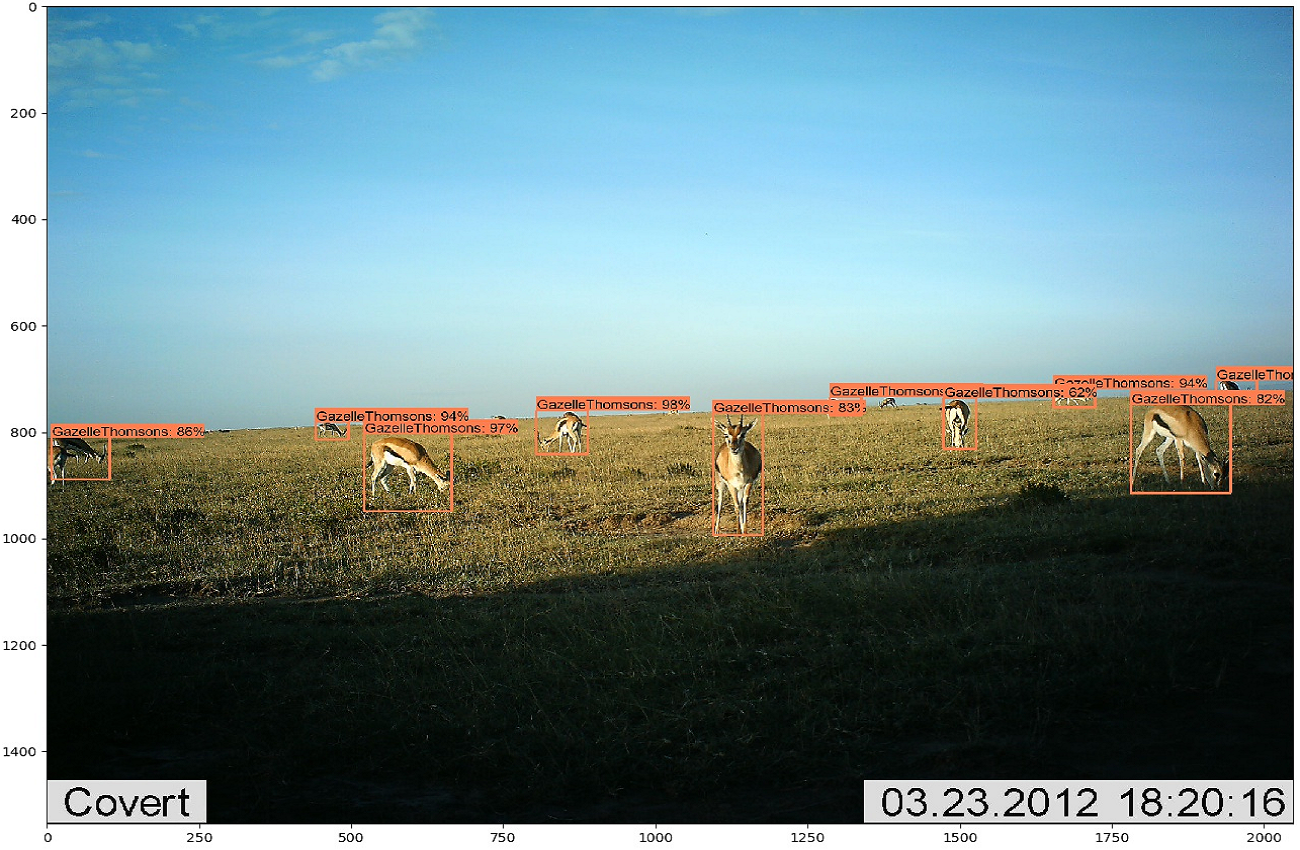}
	\caption{Faster R-CNN output returning 10 Gazelle Thomsons from the GSSS data set, demonstrating the difficulties of distances.}
\end{figure}

\onecolumn

\begin{table}[h]
\centering
\caption{Reconyx Camera Trap (RCT) data set detailed break down. Faster R-CNN returned an average accuracy of 93.0\% across all classifications. RCT contains a relatively even class distribution, likely attributing to Faster R-CNN's success. The standard deviations are quite high due to the limited number of testing images within each cross-validation set.}
\begin{tabular}{ l l c c c c }
	\multicolumn{6}{ c }{Reconyx Camera Trap}\\
	\hline
	Species & Scientific Name & \multicolumn{1}{p{1.25cm}}{\centering Total \\ Quantity} & \multicolumn{1}{p{1.25cm}}{\centering Total \\ Images} & \multicolumn{1}{p{2.5cm}}{\centering Image Class \\\ Distribution (\%)} & \multicolumn{1}{p{2cm}}{\centering Average Accuracy (\%)} \\ \hline
	Mouflon & \textit{Ovis orientalis orientalis} & 126 & 45 & 4.8 & 100.0 $\pm$ 0.0 \\
	Collared Peccary & \textit{Pecari tajacu} & 96 & 82 & 8.7 & 71.4 $\pm$ 24.4 \\
	Agouti & \textit{Dasyprocta} & 87 & 87 & 9.2 & 91.7 $\pm$ 12.5 \\
	Wild Boar & \textit{Sus scrofa} & 81 & 56 & 5.9 & 100.0 $\pm$ 0.0 \\
	Red Deer & \textit{Cervus elaphus} & 68 & 68 & 7.2 & 100.0 $\pm$ 0.0 \\
	Red Brocket Deer & \textit{Mazama americana} & 63 & 63 & 6.7 & 100.0 $\pm$ 0.0 \\
	Ocelot & \textit{Leopardus pardalis} & 63 & 63 & 6.7 & 100.0 $\pm$ 0.0 \\
	White Nosed Couti & \textit{Nasua narica} & 60 & 38 & 4.0 & 100.0 $\pm$ 0.0 \\
	Paca & \textit{Cuniculus} & 57 & 57 & 6.0 & 100.0 $\pm$ 0.0 \\
	Great Tinamou & \textit{Tinamus major} & 52 & 44 & 4.6 & 50.0 $\pm$ 28.9 \\
	White Tailed Deer & \textit{Odocoileus virginianus} & 47 & 47 & 5.0 & 100.0 $\pm$ 0.0 \\
	Roe Deer & \textit{Capreolus capreolus} & 46 & 46 & 4.9 & 100.0 $\pm$ 0.0 \\
	Common Opossum & \textit{Didelphis marsupialis} & 44 & 44 & 4.6 & 100.0 $\pm$ 0.0 \\
	Red Squirrel & \textit{Sciurus vulgaris} & 39 & 39 & 4.1 & 66.7 $\pm$ 19.2 \\
	Bird Species & \textit{Unlabeled} & 38 & 29 & 3.1 & 100.0 $\pm$ 0.0 \\
	Spiny Rat & \textit{Echimyidae} & 34 & 34 & 3.6 & 88.9 $\pm$ 19.6 \\
	European Hare & \textit{Lepus europaeus} & 31 & 28 & 3.0 & 33.3 $\pm$ 38.6 \\
	Wood Mouse & \textit{Apodemus sylvaticus} & 29 & 29 & 3.1 & 100.0 $\pm$ 0.0 \\
	Red Fox & \textit{Vulpes vulpes} & 25 & 25 & 2.6 & 100.0 $\pm$ 0.0 \\
	Coiban Agouti & \textit{Dasyprocta coibae} & 23 & 23 & 2.4 & 50.0 $\pm$ 28.6 \\

\end{tabular}
\end{table}
\newpage

\begin{table}[h]
\centering
\caption{Gold Standard Snapshot Serengeti (GSSS) data set detailed break down. GSSS contains a highly imbalanced class distribution likely related to its poor performance accuracy outside of a few main classifications. Faster R-CNN returned an average accuracy of 76.7\% across all classifications.}
\begin{tabular}{ l l c c c c }
	\multicolumn{6}{ c }{Gold Standard Snapshot Serengeti}\\
	\hline
	Species & Scientific Name & \multicolumn{1}{p{1.25cm}}{\centering Total \\ Quantity} & \multicolumn{1}{p{1.25cm}}{\centering Total \\ Images} & \multicolumn{1}{p{2.5cm}}{\centering Image Class \\\ Distribution (\%)} & \multicolumn{1}{p{2cm}}{\centering Accuracy (\%)} \\ \hline
	Wildebeest & \textit{Connochaetes} & 11321 & 1610 & 40.0 & 89.1 $\pm$ 6.2 \\
	Zebra & \textit{Equus quagga} & 3677 & 767 & 18.9 & 61.7 $\pm$ 11.2 \\
	Buffalo & \textit{Syncerus caffer} & 987 & 227 & 6.00 & 37.0 $\pm$ 28.6 \\
	Gazelle Thomsons & \textit{Eudorcas thomsonii} & 938 & 198 & 4.88 & 92.0 $\pm$ 8.3 \\
	Impala & \textit{Aepyceros melampus} & 541 & 149 & 3.67 & 66.7 $\pm$ 19.2 \\
	Hartebeest & \textit{Alcelaphus buselaphus} & 351 & 242 & 5.96 & 80.0 $\pm$ 7.0 \\
	Guineafowl & \textit{Numididae} & 195 & 54 & 1.33 & 87.5 $\pm$ 8.6 \\
	Gazelle Grants & \textit{Nanger granti} & 176 & 61 & 1.50 & 12.0 $\pm$ 6.5 \\
	Warthog & \textit{Phacochoerus africanus} & 162 & 105 & 2.59 & 33.3 $\pm$ 14.6 \\
	Elephant & \textit{Loxodonta} & 125 & 85 & 2.10 & 50.0 $\pm$ 28.9 \\
	Giraffe & \textit{Giraffa} & 121 & 87 & 2.14 & 90.0 $\pm$ 12.7 \\
	Other Bird & \textit{Unlabeled} & 77 & 48 & 1.18 & 0.0 $\pm$ 0.0 \\
	Human & \textit{Homo sapiens sapiens} & 67 & 59 & 1.45 & 60.0 $\pm$ 14.6 \\
	Stork & \textit{Ciconia ciconia} & 63 & 12 & 0.296 & 50 $\pm$ 19.1 \\
	Spotted Hyena & \textit{Crocuta crocuta} & 62 & 54 & 1.33 & 50.0 $\pm$ 38.5 \\
	Eland & \textit{Taurotragus oryx} & 48 & 24 & 0.592 & 14.6 $\pm$ 19.2 \\
	Reedbuck & \textit{Redunca} & 44 & 29 & 0.715 & 66.7 $\pm$ 34.4 \\
	Oxpecker & \textit{Buphagus} & 43 & 14 & 0.345 & 0.0 $\pm$ 0.0 \\
	Baboon & \textit{Papio} & 35 & 22 & 0.542 & 14.3 $\pm$ 14.8 \\
	Lion & \textit{Panthera leo} & 34 & 17 & 0.419 & 8.4 $\pm$ 19.2 \\
	Hippopotamus & \textit{Hippopotamus amphibius} & 32 & 28 & 0.690 & 75.0 $\pm$ 14.0 \\
	Buff Crested Bustard & \textit{Eupodotis gindiana} & 27 & 15 & 0.370 & 0.0 $\pm$ 0.0 \\
	Topi & \textit{Damaliscus korrigum} & 24 & 16 & 0.394 & 0.0 $\pm$ 0.0 \\
	Cattle Egret & \textit{Bubulcus ibis} & 86 & 15 & 1.50 & 0.0 $\pm$ 0.0 \\
	Mongoose & \textit{Herpestidae} & 11 & 5 & 0.123 & 0.0 $\pm$ 0.0 \\
	Porcupine & \textit{Hystrix africaeaustralis} & 10 & 8 & 0.197 & 0.0 $\pm$ 0.0 \\
	Kori Bustard & \textit{Ardeotis kori} & 10 & 8 & 0.197 & 0.0 $\pm$ 0.0 \\
	Cheetah & \textit{Acinonyx jubatus} & 7 & 6 & 0.148 & 0.0 $\pm$ 0.0 \\
	Dik-dik & \textit{Madoqua} & 7 & 7 & 0.173 & 0.0 $\pm$ 0.0 \\
	Superb Starling & \textit{Lamprotornis superbus} & 6 & 3 & 0.0739 & 0.0 $\pm$ 0.0 \\
	Serval & \textit{Leptailurus serval} & 6 & 6 & 0.148 & 0.0 $\pm$ 0.0 \\
	Aardvark & \textit{Orycteropus afer} & 4 & 4 & 0.986 & 0.0 $\pm$ 0.0 \\
	Secretary Bird & \textit{Sagittarius serpentarius} & 4 & 4 & 0.0986 & 0.0 $\pm$ 0.0 \\
	Leopard & \textit{Panthera pardus} & 4 & 3 & 0.0739 & 0.0 $\pm$ 0.0 \\
	Buckbuck & \textit{Tragelaphus sylvaticus} & 4 & 4 & 0.0986 & 0.0 $\pm$ 0.0 \\
	Jackal & \textit{Canis mesomelas} & 3 & 3 & 0.0739 & 0.0 $\pm$ 0.0 \\
	Other Rodent & \textit{Unlabeled} & 3 & 1 & 0.0246 & 0.0 $\pm$ 0.0 \\
	Wattled Starling & \textit{Creatophora cinerea} & 3 & 1 & 0.0246 & 0.0 $\pm$ 0.0 \\
	Aardwolf & \textit{Proteles cristata} & 2 & 2 & 0.0493 & 0.0 $\pm$ 0.0 \\
	Ostrich & \textit{Struthio camelus} & 2 & 2 & 0.0493 & 0.0 $\pm$ 0.0 \\
	Hare & \textit{Lepus microtis} & 1 & 1 & 0.0246 & 0.0 $\pm$ 0.0 \\
	Grey Backed Fiscal & \textit{Lanius excubitoroides} & 1 & 1 & 0.0246 & 0.0 $\pm$ 0.0 \\
	Rhinoceros & \textit{Rhinocerotidae} & 1 & 1 & 0.0246 & 0.0 $\pm$ 0.0 \\
	Vervet Monkey & \textit{Chlorocebus pygerythrus} & 1 & 1 & 0.0246 & 0.0 $\pm$ 0.0 \\
	Waterbuck & \textit{Kobus ellipsiprymnus} & 1 & 1 & 0.0246 & 0.0 $\pm$ 0.0 \\
	White-Headed Buffalo Weaver & \textit{Dinemellia dinemelli} & 1 & 1 & 0.0246 & 0.0 $\pm$ 0.0 \\
\end{tabular}
\end{table}
\end{document}